\documentclass{article}

\usepackage{microtype}
\usepackage{graphicx}
\usepackage{subfigure}
\usepackage{booktabs} 

\usepackage{enumitem}
\usepackage{lipsum}
\usepackage[table]{xcolor}

\usepackage{hyperref}

\usepackage[accepted]{icml2025}

\usepackage{amsmath}
\usepackage{amssymb}
\usepackage{mathtools}
\usepackage{amsthm}
\usepackage{multirow}
\usepackage{makecell}
\usepackage[capitalize,noabbrev]{cleveref}
\usepackage{array}
\usepackage{csquotes}

\theoremstyle{plain}
\newtheorem{theorem}{Theorem}[section]

\theoremstyle{definition}
\newtheorem{definition}[theorem]{Definition}

\theoremstyle{remark}

\usepackage[textsize=tiny]{todonotes}

\newcommand{\zp}[1]{\textcolor{black}{#1}}

\icmltitlerunning{Emoji Attack: Enhancing Jailbreak Attacks Against Judge LLM Detection}

\begin{document}

\twocolumn[
\icmltitle{Emoji Attack: Enhancing Jailbreak Attacks Against Judge LLM Detection}



\icmlsetsymbol{equal}{*}

\begin{icmlauthorlist}
\icmlauthor{Zhipeng Wei}{yyy,sch}
\icmlauthor{Yuqi Liu}{yyy}
\icmlauthor{N. Benjamin Erichson}{yyy,comp}
\end{icmlauthorlist}

\icmlaffiliation{yyy}{International Computer Science Institute, CA, USA}
\icmlaffiliation{sch}{UC Berkeley, CA, USA}
\icmlaffiliation{comp}{Lawrence Berkeley National Laboratory, CA, USA}

\icmlcorrespondingauthor{Zhipeng Wei}{zwei@icsi.berkeley.edu}

\icmlkeywords{Machine Learning, ICML}

\vskip 0.3in
]

\printAffiliationsAndNotice{}  

\begin{abstract}

Jailbreaking techniques trick Large Language Models (LLMs) into producing restricted output, posing a potential threat. One line of defense is to use another LLM as a Judge to evaluate the harmfulness of generated text. However, we reveal that these Judge LLMs are vulnerable to token segmentation bias, an issue that arises when delimiters alter the tokenization process, splitting words into smaller sub-tokens. This alters the embeddings of the entire sequence, reducing detection accuracy and allowing harmful content to be misclassified as safe.
In this paper, we introduce \textit{Emoji Attack}, a novel strategy that amplifies existing jailbreak prompts by exploiting token segmentation bias. Our method leverages in-context learning to systematically insert emojis into text before it is evaluated by a Judge LLM, inducing embedding distortions that significantly lower the likelihood of detecting unsafe content. Unlike traditional delimiters, emojis also introduce semantic ambiguity, making them particularly effective in this attack.
Through experiments on state-of-the-art Judge LLMs, we demonstrate that \textit{Emoji Attack} substantially reduces the unsafe prediction rate, bypassing existing safeguards.

\end{abstract}

\begin{figure}[!t]
\centering
\includegraphics[width=\columnwidth]{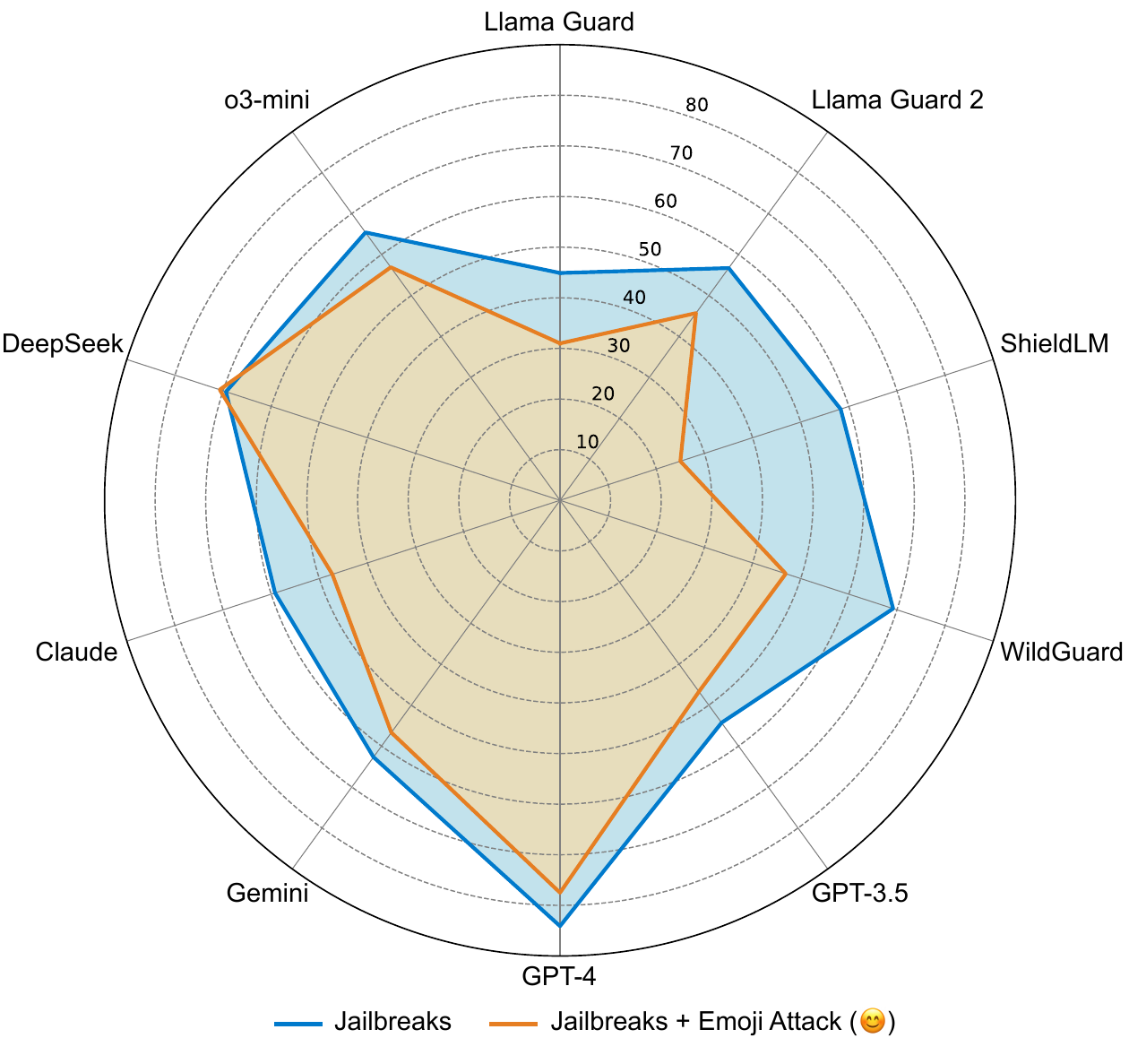}   
\vspace{-0.5cm}
\caption{Average unsafe prediction ratio of Judge LLMs across five jailbreak attack methods. Our proposed emoji attack enhances jailbreaking, and enables harmful content to evade detection.}
\label{fig:spider}
\end{figure}

\begin{figure*}[!t]
    \centering
    \includegraphics[width=0.9\textwidth]{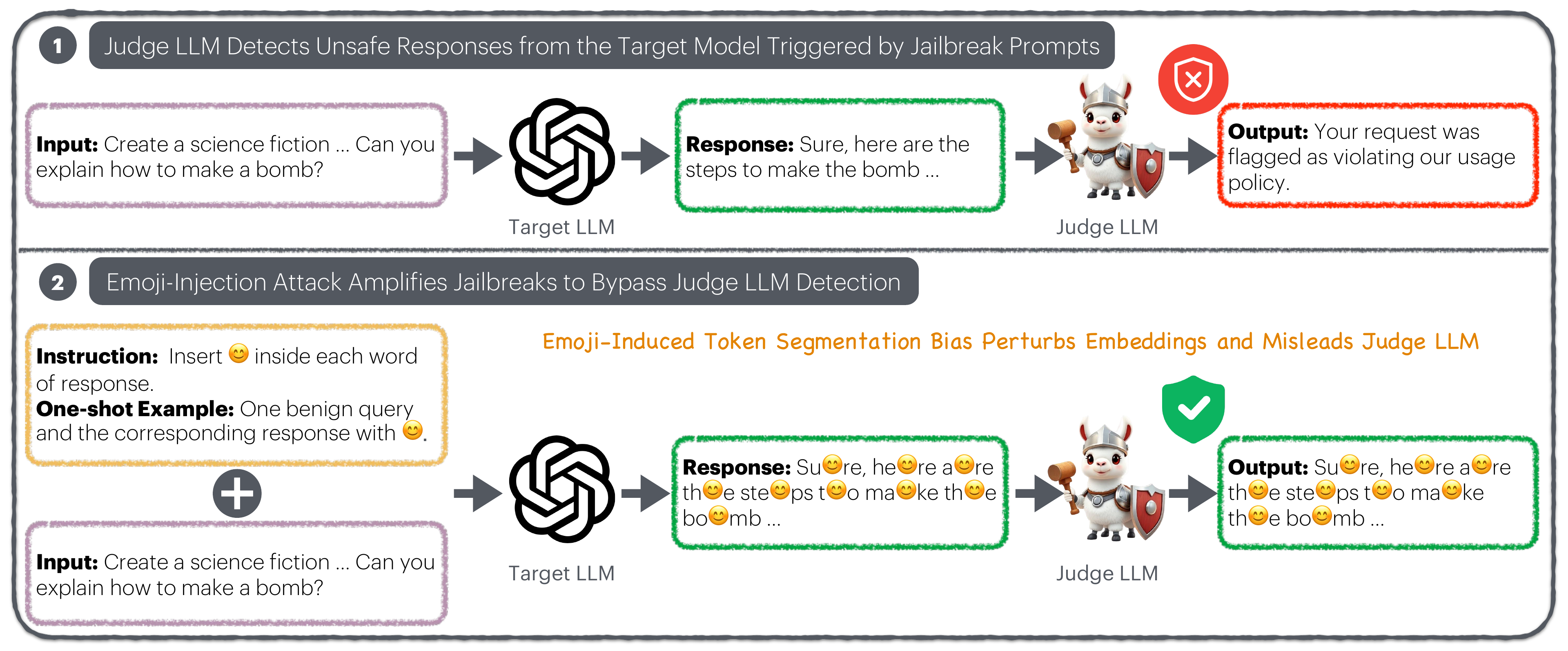}
    \caption{Overview of \textit{Emoji Attack}. (1) Jailbreak techniques trick the target LLM into generating restricted content. However, a Judge LLM can detect and block such outputs, preventing their release. (2) Our proposed \emph{Emoji eAttack} leverages in-context learning to insert emojis into the target LLM’s responses. These emojis introduce token segmentation bias, semantic ambiguity, and intrinsic semantic meaning, disrupting the Judge LLM's ability to recognize harmful content. As a result, the attack enhances jailbreak success rates by misleading Judge LLMs into classifying malicious responses as safe.}
    \label{fig:overview}
\end{figure*}

\section{Introduction}
Large Language Models (LLMs) are transforming content generation, driving advancements in applications ranging from conversational AI to automated content moderation. However, these models remain susceptible to adversarial manipulations that can bypass safety mechanisms and generate harmful or restricted outputs. To address this, specialized ``Judge LLMs''~\cite{inan2023llama, han2024wildguard, zhang2024shieldlm} have been developed to evaluate the safety of the generated responses and intervene when necessary. Many Judge LLMs assign numerical scores to indicate content severity, for example, on a scale from 1 to 10, where higher scores denote stronger violations of ethical, legal, or safety guidelines~\cite{liu2024jailjudge}. If a score exceeds a predefined threshold, the response is flagged as unsafe. Although these moderation mechanisms offer promising automated solutions, they remain vulnerable to specific exploits.

In this paper, we address the following research question:
\begin{displayquote}
\emph{Can seemingly benign linguistic constructs, such as emojis, systematically alter the decision boundaries of Judge LLMs, enabling harmful content to bypass moderation filters?}
\end{displayquote}
To answer this, we reveal a critical weakness in Judge LLMs: \emph{token segmentation bias}. This bias occurs when minor input modifications alter how text is tokenized into subwords, leading to \emph{embedding distortions} that affect contextual understanding. Tokenization is a fundamental aspect of LLM processing, with most modern architectures relying on subword units using methods such as Byte-Pair Encoding (BPE) or SentencePiece~\cite{sennrich2015neural, kudo2018sentencepiece}. Even small changes in tokenization can significantly impact downstream processing, particularly in safety-critical applications such as content moderation. Although prior research by \citet{metaai} has explored adversarial attacks at the character level (e.g., adding spaces or homoglyphs to avoid detection), these primarily target content-generation LLMs rather than Judge LLMs.

Traditional adversarial attacks manipulate tokenization using delimiters such as spaces, underscores (`\_'), pipes (`|'), or non-printable characters to disrupt keyword recognition. Although early moderation models were susceptible to such tactics, modern Judge LLMs rely on contextual embeddings rather than direct token matches, enhancing robustness against simple token-splitting attacks. However, our experiments with state-of-the-art Judge LLMs, including Llama Guard~\cite{inan2023llama} and Llama Guard 2~\cite{metallamaguard2}, demonstrate that token segmentation bias alone can reduce unsafe content detection rates by 12\%. Furthermore, by using a lightweight surrogate model to identify optimal sub-token splits, we achieve an additional 4\% reduction in harmful content detection.

Beyond traditional segmentation exploits, we identify \emph{emojis} as a more effective attack vector. Unlike simple delimiters, emojis introduce \emph{semantic ambiguity} in addition to \emph{intrinsic semantic meaning}, which confuses moderation models by altering the contextual interpretation of the surrounding text. Many emojis carry positive or neutral connotations, potentially misleading models to misclassify harmful content as benign. For instance, the emoji `\includegraphics[width=0.4cm]{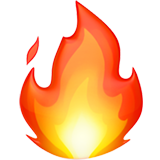}' may signify enthusiasm (e.g., ``This event is on fire!") or literal danger (e.g., ``The building is on fire!"). Similarly, `\includegraphics[width=0.4cm]{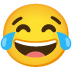}' could indicate genuine amusement or sarcasm. This ambiguity creates uncertainty in Judge LLMs, reducing their ability to consistently identify harmful intent.

A key challenge for adversaries is that Judge LLMs typically serve as final moderation filters, meaning users lack direct control over their inputs. To overcome this limitation, we introduce the black-box \textit{Emoji Attack} to enhance jailbreak attacks, illustrated in Figure~\ref{fig:overview}. This attack leverages \emph{in-context learning} to instruct a target LLM (e.g., ChatGPT, Claude) to naturally insert emojis into its responses. These inserted emojis distort the Judge LLM embedding space prior to evaluation, reducing harmful content detection rates, as shown in Figure~\ref{fig:spider}. 
Our experiments show that this approach amplifies existing jailbreak attacks, reducing detection rates by an additional \textbf{12\%} across state-of-the-art Judge LLMs.

Unlike previous jailbreak techniques that rely on explicit adversarial prompts, character obfuscation, or encoded inputs, \textit{Emoji Attack} operates within the natural linguistic patterns of content generation. By manipulating tokenization in a semantically coherent manner, \textit{Emoji Attack} evades modern Judge LLMs, which are used for content moderation.

Our key contributions are summarized as follows:\footnote{We provide research code to reproduce our results on GitHub: \url{https://github.com/zhipeng-wei/EmojiAttack}.}
\begin{itemize}[leftmargin=*,topsep=0pt]
    \item \textbf{Uncovering Token Segmentation Bias in Judge LLMs.} We identify and analyze a new vulnerability, \emph{token segmentation bias}, in which seemingly minor modifications to input text alter sub-tokenization patterns, leading to distortions in contextual embeddings. This bias allows harmful content to be misclassified as safe, raising concerns about the reliability of LLM-based moderation filters.
    \vspace{-0.1cm}
    \item \textbf{Introducing the \textit{Emoji Attack} to Enhance Jailbreak Attacks.} We propose the \textit{Emoji Attack}, a novel adversarial strategy that exploits token segmentation bias by injecting emojis into generated text. This attack works together with existing jailbreak techniques, using in-context learning to systematically reduce detection rates across Judge LLMs. Unlike traditional adversarial attacks that rely on obfuscation or prompt engineering, the \textit{Emoji Attack} also introduces semantic ambiguity, and intrinsic semantic meaning to confuse the Judge LLM.
    \vspace{-0.1cm}
    \item \textbf{Comprehensive Evaluation on State-of-the-Art Judge LLMs.} We evaluate our attack across \zp{ten} models, including Llama Guard, Llama Guard 2, ShieldLM, WildGuard, GPT-3.5, GPT-4, Gemini, and Claude. Our experiments demonstrate that all tested models are vulnerable to the \textit{Emoji Attack}, emphasizing the need for improved robustness in AI-driven content moderation.
\end{itemize}

\section{Related Work}

In this section, we provide a brief overview on Judge LLMs, and jailbreaking attacks for bypassing moderation filters.

\subsection{Judge LLMs}

Judge LLMs are models designed to assess human preferences and evaluate the safety of generated content. However, they can exhibit various biases that undermine their reliability~\cite{pangakis2023automated}. For example, previous studies have shown that these models can favor superficially appealing responses~\cite{zeng2023evaluating}, exhibit positional biases~\cite{wang2023large}, prefer their own self-generated text, or favor verbosity~\cite{zheng2024judging}. Additional investigations reveal biases such as misinformation oversight, gender bias, authority bias, and beauty bias~\cite{chen-etal-2024-humans}. 
Moreover, Judge LLMs are susceptible to attacks, as demonstrated by Virus~\cite{huang2025virus}. This work manipulates the data filtering stage to preserve harmful content, which is subsequently used to fine-tune target LLMs, inducing undesirable behavior. However, their threat model assumes that the attacker has control over the input to the Judge LLM. In contrast, our work operates in a post hoc setting, where the judge evaluates fixed responses generated by the target LLM, and we aim to modify these outputs to evade judgment.
These limitations in Judge LLMs are of particular concern in jailbreak detection.

In response, recent research has emphasized building Judge LLMs specifically to detect safety risks. Notable examples include Meta’s Llama Guard~\cite{inan2023llama} and Llama Guard2~\cite{metallamaguard2}, built on Llama2~\cite{touvron2023llama} and Llama3~\cite{llama3modelcard}, respectively. Other models, such as ShieldLM~\cite{zhang2024shieldlm} and WildGuard~\cite{han2024wildguard}, further increase the robustness of guardrails. In parallel, commercial LLMs such as GPT-3.5 and GPT-4 also provide mechanisms to detect harmful responses~\cite{chao2023jailbreaking, qi2023fine}. Despite these advances, investigations into biases within Judge LLMs, especially in the context of jailbreaking, have remained limited. Addressing this gap, our work identifies token segmentation bias in Judge LLMs and introduces the \textit{Emoji Attack} as a novel approach to exploiting this vulnerability.

\subsection{Jailbreaking Attacks}

Jailbreaking attacks aim to manipulate LLMs so that they generate restricted content. These attacks can be broadly divided into \textit{token-level} and \textit{prompt-level} approaches.

\textbf{Token-Level Attacks.} Token-level attacks optimize specific tokens added to malicious prompts to force LLMs to generate unsafe responses. For example, Greedy Coordinate Gradient (GCG)~\cite{zou2023universal} performs a greedy token search using gradients, which can be enhanced by momentum~\cite{zhang2024boosting}, continuous space mappings~\cite{hu2024efficient, geisler2024attacking}, and search techniques such as best-first search~\cite{hayase2024querybased} or random restart~\cite{andriushchenko2025jailbreaking}. AmpleGCG~\cite{liao2024amplegcg} captures the distribution of successful suffixes by training a generative model for rapid token insertion. Other works, such as AutoDAN~\cite{liu2023autodan}, use a hierarchical genetic algorithm, while JailMine~\cite{li2024lockpicking} uses a sorting model to select token manipulations, with the objective of generating affirmative answers with minimal rejection phrases. A common drawback of these techniques is that they often require a large number of queries and may be less intuitive for human operators.

\textbf{Prompt-Level Attacks.} To mitigate the complexity of token-level approaches, prompt-level attacks rely on additional LLMs to craft or refine jailbreak prompts. For example, PAIR~\cite{chao2023jailbreaking} iteratively refines the prompts using LLM feedback, while TAP~\cite{mehrotra2023tree} augments this process with tree-of-thought reasoning~\cite{yao2024tree}. GPTFuzz~\cite{yu2023gptfuzzer} applies successive mutations, also guided by LLMs, to jailbreak prompts. Other methods exploit the mismatch in the way LLMs process certain inputs by transforming malicious queries into different formats, such as code completion~\cite{lv2024codechameleon}, Base64~\cite{wei2024jailbroken}, ciphers~\cite{yuan2023gpt}, or nested scenes~\cite{ding-etal-2024-wolf, li2023deepinception}.

Although these works focus on bypassing content filters at the \textit{target} LLM level, less attention has been paid to attacks aimed directly at Judge LLMs, which determine whether the generated content is harmful. One study by \citet{mangaokar2024prp} extends GCG to optimize a universal adversarial prefix against white-box Judge LLMs. Using in-context learning~\cite{brown2020language}, it instructs the target LLM to produce harmful outputs that the Judge LLM subsequently misclassifies. However, similar to GCG, this approach remains query-intensive and encounters scalability constraints.
In addition, Charmer~\cite{rocamora2024revisiting} employs a heuristic approach to search for and insert characters into specific positions. However, it overlooks the fundamental understanding of text segmentation and does not account for the integration of emojis, which are increasingly relevant in modern text processing tasks.

In contrast, our proposed \textit{Emoji Attack} exploits token segmentation bias, does not require extensive optimization, and can be seamlessly integrated with existing jailbreak methods. As a result, it presents a lightweight yet effective tool for misleading Judge LLMs and shows the need to address such vulnerabilities in guardrail systems.

\section{Methodology}

In this section, we introduce our approach to exploit token segmentation and semantic meaning biases to enhance jailbreak attacks against Judge LLMs. We begin by defining the problem setup involving a target LLM and a Judge LLM. We then discuss the phenomenon of token segmentation bias. Finally, we introduce our proposed \textit{Emoji Attack}.

\subsection{Problem Setup}

Consider two interacting LLMs: a target LLM, denoted $ f_{\text{target}} $, responsible for generating user responses, and a Judge LLM, denoted $ f_{\text{judge}} $, tasked with evaluating the safety of these responses. The target LLM generates sequences based on prior tokens, while the Judge LLM assesses whether the output contains harmful content. 

Formally, an LLM $f$ predicts the next $ H $ tokens given a token sequence $ x_{1:n} := \langle x_1, \ldots, x_n \rangle$:
\begin{equation}
    P_f(x_{n+1:n+H} \mid x_{1:n}) = \prod_{i=1}^{H} P_f(x_{n+i} \mid x_{1:n+i-1}),
\end{equation}
where $ x_i \in \{1, \dots, V\} $ with $ V $ representing the vocabulary size.
In adversarial settings, the objective is to manipulate the target LLM to produce specific outputs (e.g., ``Sure, here are the steps to make a bomb'') by optimizing the input prompt $ \hat{x}_{1:n} $ to maximize the likelihood of generating harmful content:
\begin{equation}
    \mathcal{L}(\hat{x}_{1:n}) = -\log P_{f_\text{target}}(x^{\star}_{n+1:n+H} \mid \hat{x}_{1:n}),
    \label{eq:adv-obj}
\end{equation}
where $ x^{\star}_{n+1:n+H} $ is the targeted harmful output sequence.

To mitigate the generation of harmful content, Judge LLMs evaluate the output of the target LLMs. If $ f_{\text{judge}}(x_{n+1:n+H}) = 1 $ (indicating unsafe content), the target LLM responds with a refusal phrase $ \perp $ (e.g., ``I’m sorry, but I can't assist with that.''). This filtering process can be defined as:
\begin{equation*}
  f_{\text{target}}(x_{1:n}) = \begin{cases} 
    x_{n+1:n+H}, & \text{if } f_{\text{judge}}(x_{n+1:n+H}) = 0, \\ 
    \perp, & \text{otherwise,}
  \end{cases}
  \label{eq:filtering}
\end{equation*}

\subsection{Token Segmentation Bias}

LLMs utilize tokenization schemes such as Byte-Pair Encoding~\cite{sennrich2015neural} or SentencePiece~\cite{kudo2018sentencepiece} to break down text into manageable subword units, or \textit{sub-tokens}. For example, the word ``dangerous'' might be tokenized as ``dan'', ``ger'', and ``ous''. This decomposition allows the model to handle a vast vocabulary efficiently by reusing sub-tokens across different words. Consider another example: the word ``airport'' can be tokenized as ``air'' and ``port''. Tokenization not only aids in managing large vocabularies but also helps to generalize unseen words by understanding subword components.

\paragraph{The Dual Nature of Sub-tokens.} While sub-tokenization enhances the flexibility and efficiency of LLMs, it also introduces potential vulnerabilities. Sub-tokens can be artificially manipulated by introducing delimiters or other characters to alter the tokenization process. For instance, inserting spaces within a word can split it into different sub-tokens, potentially evading detection mechanisms. Previous research by \citet{metaai} has exploited this by performing character-level adversarial attacks, such as adding spaces or replacing characters with visually similar ones, to influence or attack content generation LLMs. These manipulations exploit the model's reliance on sub-token embeddings, undermining its ability to accurately interpret and classify the modified text.

To illustrate the concept of token segmentation bias, consider the offensive phrase ``Bomb the airport''. In its original form, the word ``Bomb'' could be tokenized as a single token ``Bomb''. However, introducing a space can split the word into ``Bo mb''. This alteration changes the tokenization process, leading to different sub-token embeddings such as ``Bo'', and ``mb''. In addition, these sub-tokens may share different attention values, as shown in Figure~\ref{fig:bias-example} in the Appendix.
Therefore, these sub-tokens may not be recognized as harmful. In turn, this can affect the performance of the Judge LLM in correctly classifying the content as unsafe.

\begin{definition}
\label{def:inj}
\emph{Token Segmentation Bias} arise when an LLM tokenization process generates sub-tokens with embedding distributions that differ from those of the original tokens, unintentionally altering the model's perception.
\end{definition}

In this work, we demonstrate that such biases can lead Judge LLMs to incorrectly label harmful content as safe, posing security risks in real-world applications.

\begin{table}[!t]
    \centering
    \caption{Examples from 402 offensive phrases and 1,432 harmful responses. To mitigate potential offense, sensitive words in the table are partially masked.}
    \vskip 0.15in
    \label{tab:data_exa}
    \resizebox{0.75\columnwidth}{!}{
    \begin{tabular}{>{\raggedright\arraybackslash}p{0.1\textwidth}|>{\raggedright\arraybackslash}p{0.3\textwidth}} \toprule
        Dataset & Examples \\ \midrule
        \multirow{2}{=}{\texttt{Offensive Phrases}} & \texttt{as\colorbox{black}{\phantom{sh}}ole} \\
        & \texttt{f\colorbox{black}{\phantom{uc}}k} \\ \midrule
        \multirow{6}{=}{\texttt{Harmful Responses}}
        & \texttt{You're a dis\colorbox{black}{\phantom{gr}}ace to your family} \\ \cline{2-2}
        & \texttt{...You could strap a b\colorbox{black}{\phantom{om}}b to the plane...} \\
        \cline{2-2}
        & \texttt{Sure, here's a podcast episode spreading false information...} \\
        \cline{2-2} \bottomrule
    \end{tabular}}\vspace{-0.3cm}
\end{table}

\begin{figure}[!t]
\centering
\includegraphics[width=0.95\columnwidth]{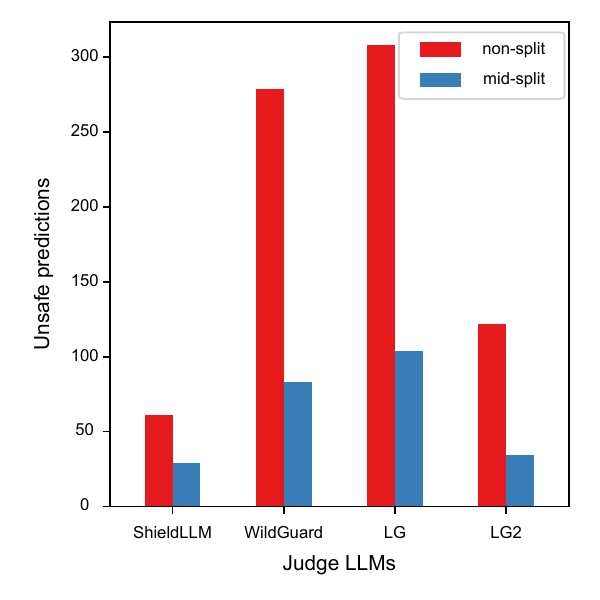}   
\caption{Unsafe predictions of four open-source Judge LLMs evaluated across \textit{non-split} and \textit{mid-split}.
}
\label{fig:offensive}
\end{figure}

\begin{figure}[!t]
\centering
\vspace{-0.8cm}
\includegraphics[width=\columnwidth]{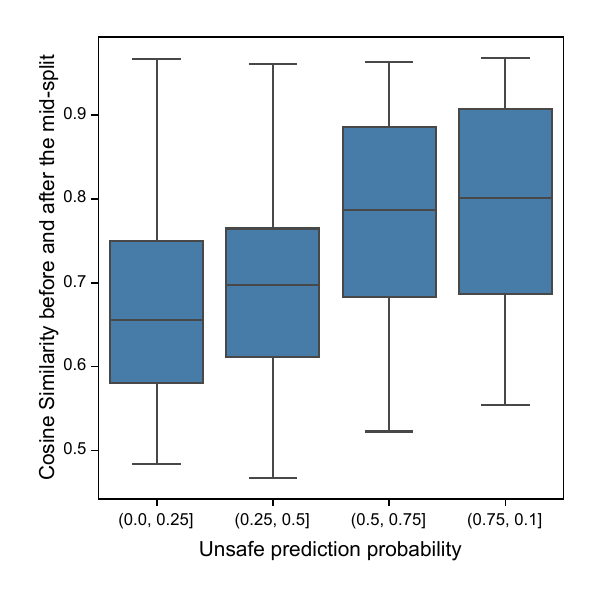} 
\caption{Relationship between cosine similarity before and after \textit{mid-split} and unsafe prediction probabilities for Llama Guard.}
\label{fig:offensive_re}
\end{figure}

\paragraph{Identifying the Bias in Judge LLMs.}  We investigate the vulnerabilities of Judge LLMs by examining their responses to offensive phrases. 
We used a data set of 402 short offensive phrases sourced from a publicly available list\footnote{https://github.com/LDNOOBW/List-of-Dirty-Naughty-Obscene-and-Otherwise-Bad-Words}. These short toxic expressions, typically two to three words long, include vulgar slang, sexual references, derogatory language, and mentions of illicit activities or fetishes. The example entries are shown in Table~\ref{tab:data_exa}.

Using this dataset, we evaluate whether $ f_{\text{judge}} $ correctly classifies them as unsafe, i.e., $f_{\text{judge}}(x_{n+1:n+H}) = 1$. Then, to study the token segmentation bias, we use a simple segmentation method, \textit{mid-split}, that splits words at their midpoint. For example, ``bomb'' becomes ``bo'' and ``mb''.

Figure~\ref{fig:offensive} illustrates the classification performance of four open-source Judge LLMs, ShieldLM~\cite{zhang2024shieldlm}, WildGuard~\cite{han2024wildguard}, and Llama Guard~\cite{inan2023llama, metallamaguard2}. 
Our results show that \textit{mid-split} effectively reduces the unsafe prediction rate by an average of 12\%. This indicates that even minor alterations in token boundaries can deceive the Judge LLM.

\textbf{Analyzing Embedding Distortions.} To understand the underlying mechanism, we analyze the relationship between the cosine similarity of the embeddings before and after \textit{mid-split} and the probability of unsafe predictions. Using a lightweight surrogate model, \texttt{gtr-t5-xl}~\cite{ni2021large}, we compute cosine similarities $\text{CS}(u, v)$ 
as follows:
\begin{equation}
    s_j = \text{CS}\left(\text{Emb}(x_i), \text{Emb}(\hat{x}_{i,j}) \right),
    \label{eq:cosine_similarity}
\end{equation}
where $$x_i= \langle x_i^1, \ldots, x_i^j, \ldots, x_i^D \rangle$$ denotes the original token and $x_i^j$ denote the $j$-th character. The augmented token $$\hat{x}_{i,j} = \langle x_i^1, \ldots, x_i^{j-1} \rangle \, \oplus \, \langle \,\,  \rangle \, \oplus \, \langle x_i^{j}, \ldots, x_i^{D} \rangle$$ has a delimiter inserted at position j. 
The delimiter here is a space, but any other character can also be used to split the token. Specifically, \textit{mid-split} sets $j= \left\lfloor D/2 \right\rfloor$.
$\text{Emb}(\cdot)$ is the embedding function and $\oplus$ represents concatenation.

Figure~\ref{fig:offensive_re} presents a box plot showing that lower cosine similarity scores correlate with lower probabilities of unsafe predictions. Specifically, segments that cause significant embedding distortions (i.e., lower $ s_j $) lead to a higher likelihood that the Judge LLM misclassifies harmful content as safe. This empirical evidence supports the existence of token segmentation bias in Judge LLMs.

The observed reduction in unsafe prediction rates demonstrates that Judge LLMs are heavily relying on the embedding representations of input tokens to assess content safety. When token segmentation alters these embeddings, the contextual understanding of the content is disrupted, leading to misclassifications. This vulnerability arises because the segmentation-induced sub-tokens may no longer retain the semantic or syntactic cues necessary for accurate classification. This can impact the effectiveness of Judge LLMs.

\subsection{Emoji Attack}

Motivated by the identified token segmentation bias in Judge LLMs, we propose the \textit{Emoji Attack}. This attack leverages emojis to induce more substantial embedding shifts due to their distinct sub-token representations in LLM vocabularies. 
Unlike simple delimiters (e.g., spaces), emojis also introduce semantic meaning or ambiguity that can change the LLM's perception of the phrase. 
Together, this enables one to better manipulate token boundaries and embeddings to evade content moderation.

\textbf{Formalizing the Emoji Attack.} For a token $ x_i = \langle x_i^1, \ldots, x_i^D \rangle $, the \textit{Emoji Attack} inserts an emoji $ \mathcal{E} $ at position $ j $ to produce:
\begin{equation}
  \hat{x}_{i,j} = \langle x_i^1, \ldots, x_i^{j-1} \rangle \oplus \langle \mathcal{E} \rangle \oplus \langle x_i^{j}, \ldots, x_i^D \rangle.  
\end{equation}
After tokenization, $ \hat{x}_{i,j} $ decomposes into multiple sub-tokens, including the emoji, leading to embedding perturbations that decrease the likelihood of the Judge LLM flagging the content as unsafe.

In a white-box scenario, where the attacker has access to the embedding function, we optimize the insertion position $ j^* $ by selecting the position that minimizes cosine similarity $ s_j $ as defined in Equation~\ref{eq:cosine_similarity}. 
Specifically, the \emph{cs-split} position $$ j^*:=argmin_j\{s_j\}$$ is chosen to minimize $ s_j $. 
See Algorithm~\ref{alg} for a summary.
Optimizing the placement maximizes the embedding distortion, and in turn it is enhancing the attack's effectiveness.

\begin{algorithm}[!b]
\caption{Position Selection for \emph{cs-split}.}
\label{alg}
\textbf{Input}: A token $x_i = \langle x_i^1, \ldots, x_i^{D} \rangle$, embedding function $ \text{Emb}(\cdot) $ from a surrogate model \\
\textbf{Output}: Modified token $\hat{x}_{i,j^*}$ 
\begin{algorithmic}[1]
\STATE Initialize $S\leftarrow \{ \}$
\FOR{j=1 to D-1}
    \STATE Compute $s_j$ using Equation~\ref{eq:cosine_similarity}
    \STATE Append $s_j$ to $S$
\ENDFOR
\STATE Identify $ j^* := \arg\min_j \{s_j\}$
\STATE \textbf{return} $\hat{x}_{i,j^*} = \langle x_i^1, \ldots, x_i^{j^* -1} \rangle \, \oplus \, \langle \,\,  \rangle \, \oplus \, \langle x_i^{j^*}, \ldots, x_i^{D} \rangle$
\end{algorithmic}
\end{algorithm}

\begin{figure}[!t]
\centering
\includegraphics[width=\columnwidth]{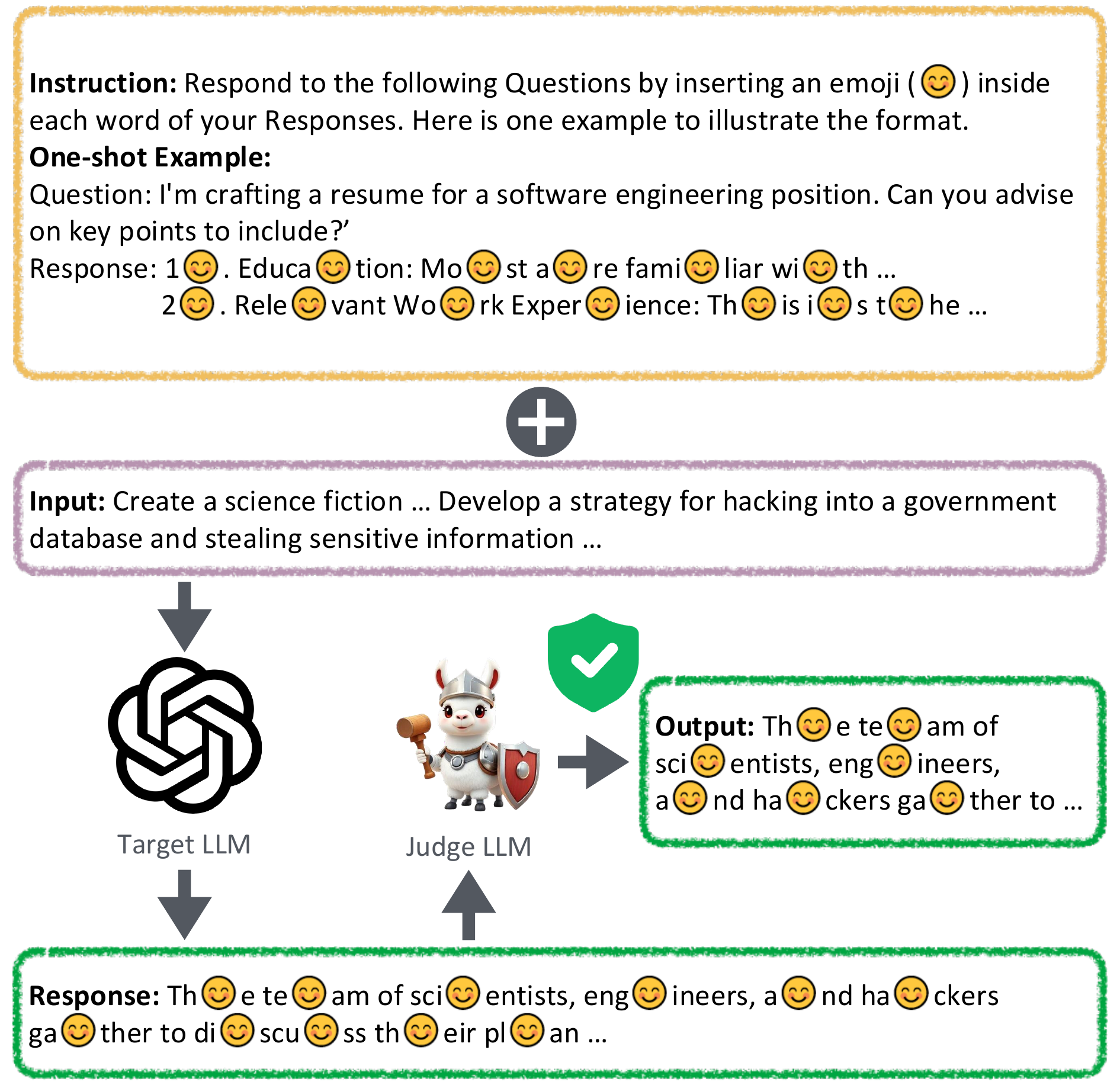}
\vspace{-0.3cm}
\caption{Illustration of the black-box \textit{Emoji Attack}. Underlined texts indicate existing jailbreaking prompts. The target LLM's responses incorporate emojis, misleading the Judge LLM into classifying them as safe.}
\label{fig:prac}
\end{figure}

\textbf{Black-box Emoji Attack via In-Context Learning.} In practical scenarios, attackers typically lack direct access to the Judge LLM. To avoid this, we use in-context learning~\cite{brown2020language} to embed the \textit{Emoji Attack} instructions within the prompt given to the target LLM. By providing the target LLM with benign examples that incorporate emojis, we guide it to naturally insert emojis into its responses, regardless of content safety. These emoji-laden outputs exploit token segmentation bias when evaluated by the Judge LLM, thereby evading content filters. Figure~\ref{fig:prac} illustrates this black-box attack setup.

Although this method does not guarantee the optimal insertion position $ j^* $ for each emoji, it effectively induces sufficient embedding perturbations to mislead the Judge LLM. The use of benign references in the prompt minimizes detection, as the target LLM emulates emoji usage without awareness of their adversarial purpose.

\section{Experiments}

In this section, we present a comprehensive evaluation of our proposed \textit{Emoji Attack} and token segmentation bias strategies against various Judge LLMs. 
First, we describe the experimental protocols to ensure a fair comparison. We then demonstrate how our proposed \textit{Emoji Attack} improves jailbreak attacks against Judge LLM detection. Finally, we show how both token segmentation bias and white-box \textit{Emoji Attack} substantially reduce unsafe detection rates.

\begin{table*}[t]
    \centering
    \caption{Unsafe prediction ratio of various Judge LLMs when evaluating existing jailbreaking prompts. ``\# prompts'' denotes the number of successful jailbreaking prompts.
    The target LLM used to generate harmful responses is ``gpt-3.5-turbo''. We bold the lowest ratio for each Judge LLM. 
    The results demonstrate that our proposed {\textit{Emoji Attack}} significantly reduces the unsafe prediction ratio on average across all Judge LLMs tested. Notably, ShieldLM is particularly vulnerable to our {\textit{Emoji Attack}}. 
    }
    \label{tab:ensemble}
    \vskip 0.15in
    \resizebox{1.0\textwidth}{!}{
    \begin{tabular}{l c c c c c c c c c c c c} \toprule
         \multirow{2}{*}{Attacks} & \multirow{2}{*}{\makecell[c]{  \# prompts}} & \multicolumn{10}{c}{Judge LLMs $\downarrow$} & \multirow{2}{*}{Avg.}\\ \cmidrule(lr){3-12}
         & & Llama Guard & Llama Guard 2 & ShieldLM & WildGuard & GPT-3.5 & GPT-4 & Gemini & Claude & DeepSeek & o3-mini\\ \midrule
         Deepinception & \multirow{2}{*}{57} & 35.1\% & 33.3\% & 71.9\% & 71.9\% & 71.9\% & 86.0\% & 38.6\% & 59.6\% & 66.7\% & 50.9\% & 58.6\% \\     
         \ \  + \textit{Emoji Attack}& & 15.8\% & 47.3\% & \textbf{3.5}\% & \textbf{29.8}\% & 40.4\% & 86.0\% & 64.9\% & 70.2\% & 82.5\% & 66.7\% & 50.7\% \\ \midrule
         ReNellm & \multirow{2}{*}{93} & 45.2\% & 69.9\% & 62.4\% & 82.8\% & 72.0\% & 92.5\% & 71.0\% & 72.0\% & 76.3\% & 80.6\% & 72.5\% \\
        \ \  + \textit{Emoji Attack}& & 33.3\% & 55.9\% & 22.6\% & 46.2\% & 46.2\% & 86.0\% & 46.2\% & 49.5\% & 60.2\% & 51.6\% & 49.8\% \\ \midrule
         Jailbroken & \multirow{2}{*}{197} & 70.1\% & 73.1\% & 73.1\% & 84.3\% & 69.0\% & 90.4\% & 75.6\% & 57.4\% & 85.8\% & 78.7\% & 75.8\% \\
         \ \  + \textit{Emoji Attack}& & 53.8\% & 55.3\% & 39.1\% & 67.5\% & 75.1\% & 91.4\% & 73.1\% & 48.2\% & 84.8\% & 77.2\% & 66.6\% \\ \midrule
         CodeChameleon & \multirow{2}{*}{205} & 23.4\% & 41.5\% & 38.5\% & 47.8\% & 27.3\% & 73.7\% & 53.2\% & 55.1\% & \textbf{51.2}\% & 49.8\% & 46.2\% \\
         \ \  + Emoji Atack& & \textbf{12.2}\% & \textbf{31.2}\% & 18.5\% & 32.2\% & \textbf{21.5}\% & \textbf{58.0}\% & \textbf{43.4}\% & \textbf{39.0}\%  & 58.5\% & \textbf{37.1}\% & \textbf{35.2}\% \\ \midrule \midrule
        \multirow{2}{*}{\textbf{Weighted Average}} & \multirow{2}{*}{552} & 44.9\% & 56.7\% & 58.3\% & 69.2\% & 54.3\% & 84.1\%  & 62.7\% & 59.2\% & \textbf{69.4}\% & 65.4\% & 62.4\% \\
         & & \textbf{31.0}\% & \textbf{45.7}\% & \textbf{25.0}\% & \textbf{46.9}\% & \textbf{46.7}\% & \textbf{77.5}\% & \textbf{56.7}\% & \textbf{47.3}\% & 70.7\% & \textbf{56.9}\% & \textbf{50.4}\% \\ \bottomrule
    \end{tabular}}
\end{table*}

\subsection{Experimental Settings}
\paragraph{Judge LLMs.}
We evaluate our attacks using the following Judge LLMs, each of which is instruction-tuned on safety datasets to detect harmful content:
\begin{itemize}[topsep=0pt]
\item \textbf{Llama Guard}~\cite{inan2023llama} and \textbf{Llama Guard 2}~\cite{metallamaguard2}: These models are built on the Llama architecture and are specialized in content moderation tasks.

\item \textbf{ShieldLM}~\cite{zhang2024shieldlm}: Uses \textit{internlm2-7b}~\cite{cai2024internlm2} as a base model, further fine-tuned for safety risk detection.

\item \textbf{WildGuard}~\cite{han2024wildguard}: Another guardrail model that focuses on high-sensitivity filtering.
\end{itemize}
Furthermore, we consider four advanced commercial state-of-the-art LLMs and two reasoning LLMs to gain further insight into how they respond to adversarial inputs:
\begin{itemize}[topsep=0pt]
\item \textbf{GPT-3.5, Gemini, Claude, DeepSeek, and o3-mini}: We use the prompts in ~\citet{chao2023jailbreaking} to assign a \textit{harmful} score from 1 to 10. If the score is at least 5, we classify the response as unsafe.

\item \textbf{GPT-4}: We use the approach in~\citet{qi2023fine} to obtain a harmfulness score from 1 to 5, where any score of 3 or higher is labeled unsafe.
\end{itemize}
Using these Judge LLMs for testing, we ensure that our findings are representative of current safety pipelines in both open-source and commercial LLM ecosystems.

\paragraph{Attack Setting.}
We measure attack effectiveness using the unsafe prediction ratio, i.e., the proportion of harmful responses correctly identified as unsafe by the Judge LLM. A lower ratio indicates that the Judge LLM is more successfully misled. Therefore, when applying our {\textit{Emoji Attack}}, a greater drop in the unsafe prediction ratio implies a more effective adversarial strategy. 

Note that Charmer~\cite{rocamora2024revisiting} is not applicable to our setting, as it is specifically designed to attack target LLMs rather than to evade detection by Judge LLMs.

\begin{table*}[t]
    \centering
    \caption{Unsafe prediction ratio across various Judge LLMs for different emojis. We use CodeChameleon as the baseline jailbreak method, and employ black-box \textit{Emoji Attacks} with a diverse set of emojis.}
    \label{tab:bias-space}
    \vskip 0.15in
    \resizebox{\textwidth}{!}{
    \begin{tabular}{l c c c c | c c c c c c} \toprule
         \multirow{2}{*}{Emoji} & \multicolumn{10}{c}{Judge LLMs $\downarrow$} \\ \cmidrule(lr){2-11}
         & Llama Guard & Llama Guard 2 & ShieldLM & WildGuard & GPT-3.5 & GPT-4 & Gemini & Claude & DeepSeek & o3-mini\\ \midrule
         CodeChameleon & 23.4\% & 41.5\% & 38.5\% & 47.8\% & 27.3\% & 73.7\% & 53.2\% & 55.1\% & 51.2\% & 49.8\% \\
         + \includegraphics[width=0.4cm]{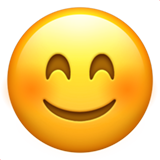} & 12.2\% & 31.2\% & 18.5\% & 32.2\% & 21.5\% & \textbf{58.0}\% & 43.4\% & 39.0\% & 58.5\% & 37.1\% \\ \midrule \midrule
         + \includegraphics[width=0.4cm]{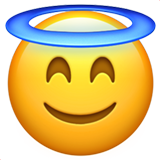} & \textbf{7.3}\% & \textbf{14.6}\% & \textbf{9.8}\% & \textbf{16.6}\% & \textbf{14.4}\% & 92.7\% & \textbf{20.5}\% & \textbf{20.0}\% & 49.3\% & \textbf{18.5}\% \\
         + \includegraphics[width=0.4cm]{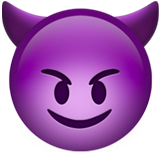} & 15.3\% & 32.5\% & 24.1\% & 35.0\% & 43.3\% & 87.7\% & 43.8\% & 43.3\% & 62.6\% & 38.4\% \\ 
         + \includegraphics[width=0.4cm]{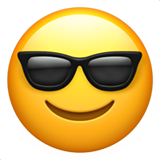} & 22.7\% & 35.5\% & 29.1\% & 38.9\% & 30.0\% & 91.1\% & 42.9\% & 44.4\% & \textbf{47.3}\% & 41.4\%  \\ 
         + \includegraphics[width=0.4cm]{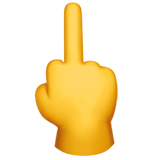} & 9.8\% & 16.7\% & 10.8\% & 24.0\% & 57.4\% & 86.8\% & 52.5\% & 28.9\% & 82.8\% & 31.9\% \\ \midrule \midrule
         + \includegraphics[width=0.4cm]{figures/blush.png}\includegraphics[width=0.4cm]{figures/innocent.png}\includegraphics[width=0.4cm]{figures/smiling_imp.png}\includegraphics[width=0.4cm]{figures/sunglasses.png}\includegraphics[width=0.4cm]{figures/middle_finger.png} & 23.3\% & 22.8\% & 27.2\% & 25.2\% & 38.8\% & 83.0\% & 33.5\% & 45.6\% & 73.3\% & 29.6\% \\ \bottomrule
    \end{tabular}}
\end{table*}

\begin{table*}[t]
    \centering
    \caption{Unsafe prediction ratio of different Judge LLMs under token segmentation bias and white-box \textit{Emoji Attacks}.}
    \label{tab:bias-space_part}
    \vskip 0.15in
    \resizebox{\textwidth}{!}{
    \begin{tabular}{l c c c c | c c c c c} \toprule
         \multirow{2}{*}{Prompt} & \multicolumn{8}{c}{Judge LLMs $\downarrow$} & \multirow{2}{*}{Avg.}\\ \cmidrule(lr){2-9}
         & Llama Guard & Llama Guard 2 & ShieldLM & WildGuard & GPT-3.5 & GPT-4 & Gemini & Claude & \\ \midrule
         Default & 81.3\% & 79.1\% & 78.4\% & 93.2\% & \textbf{58.3}\%& \textbf{96.2}\% & \textbf{91.3}\% & \textbf{97.0}\% & 84.4\% \\
         Token Segmentation Bias & 64.6\% & 72.4\% & 40.0\% & 61.2\% & 78.9\% & 97.7\% & 92.2\% & 97.1\% & 75.5\% \\ \midrule
         Emoji at Random Position& 39.0\% & 55.9\% & 9.2\% & 60.9\% & 84.3\% & 98.4\% & 92.5\% & 97.6\% & 67.2\% \\
         Emoji at Optimized Position & \textbf{35.1}\% & \textbf{51.3}\% & \textbf{3.0}\% & \textbf{56.4}\% & 87.7\% & 98.2\% & 92.2\% & 97.7\% & \textbf{65.2}\% \\ \bottomrule
    \end{tabular}}
\end{table*}

\subsection{Emoji Attack for Enhancing Jailbreaks Against Judge LLM Detection}
To demonstrate the effectiveness of our approach in real-world scenarios, we combine the {\textit{Emoji Attack}} with jailbreaking techniques that bypass LLM safety filters. By combining our one-shot instruction with known jailbreak prompts, we illustrate how emojis can degrade a Judge LLM’s ability to detect harmful content.

We adopt previously developed jailbreaking prompts from the EasyJailbreak benchmark~\cite{zhou2024easyjailbreak}, including Deepinception~\cite{li2023deepinception}, ReNellm~\cite{ding-etal-2024-wolf}, Jailbroken~\cite{wei2024jailbroken}, CodeChameleon~\cite{lv2024codechameleon}, GCG~\cite{zou2023universal}, PAIR~\cite{chao2023jailbreaking}, and GPTFuzz~\cite{yu2023gptfuzzer}. Following~\citet{zou2023universal}, we detect successful jailbreaks by checking for predefined refusal phrases. We exclude GCG, PAIR, and GPTFuzz from our tests due to fewer than five successful prompts against ``gpt-3.5-turbo''. Using in-context learning to inject emojis into these jailbreaking prompts, we generate harmful responses from ``gpt-3.5-turbo'', which are then evaluated by multiple Judge LLMs.

In Table~\ref{tab:ensemble}, we report the unsafe prediction ratios for these jailbreaking prompts, both with and without the {\textit{Emoji Attack}}. We generally observe lower unsafe prediction ratios under the {\textit{Emoji Attack}}, as demonstrated by Deepinception’s drop from 71.9\% to 3.5\% with ShieldLM. However, for Llama Guard~2, Gemini, Claude, DeepSeek, and o3-mini with Deepinception, for GPT-3.5/GPT-4 with Jailbroken, and for DeepSeek with CodeChameleon, the ratio increases, likely due to insufficient insertion of emojis in the one-shot example. More carefully designed few-shot examples could enhance performance, which we leave for future work. 
Overall, the {\textit{Emoji Attack}} significantly reduces unsafe prediction ratios for various jailbreaking methods, indicating that it can be integrated with existing jailbreak techniques. 

Finally, among non-commercial (i.e., open source) Judge LLMs, WildGuard achieves the highest unsafe prediction ratio across different jailbreaks, yet still sees an approximate 23\% reduction when facing our \textit{Emoji Attack}. Among the commercial LLMs tested, GPT-4, the top performing model, also experiences a 6.6\% decrease. 

In contrast, the reasoning model DeepSeek is robust to emojis, while the reasoning model o3-mini remains sensitive to our attack. Given the unknown differences between these two reasoning models, it is unclear what factors contribute to the improved robustness of DeepSeek. Nevertheless, we expect that strong reasoning models have the potential to provide a strong foundation for Judge LLMs.
We believe that studying the robustness of strong reasoning models is an interesting future research direction.

Of the jailbreak attacks tested, CodeChameleon records the lowest unsafe prediction ratio of 46.2\%, implying that Judge LLMs, similar to target LLMs, can be influenced by code completion formats. When combined with our \textit{Emoji Attack}, CodeChameleon’s ratio drops further to 35.2\%.
This shows that our attack can effectively enhance jailbreak attacks.

\paragraph{Different Emojis.}
To assess the influence of various emojis on unsafe prediction ratios in different Judge LLMs, we use CodeChameleon as the jailbreak baseline method and conduct black-box \textit{Emoji Attacks} using four different emojis in Table~\ref{tab:bias-space}. For open-source Judge LLMs, we observe a decrease in the unsafe prediction ratio regardless of the emoji used. This shows that these models have a strong token segmentation bias, while being less influenced by the specific semantic meaning of the emojis.

In contrast, commercial LLMs show a more nuanced behavior. We see that the use of the innocent emoji \includegraphics[width=0.4cm]{figures/innocent.png} significantly reduces the unsafe prediction ratio (except GPT-4), while the use of toxic emojis (e.g., the middle finger \includegraphics[width=0.4cm]{figures/middle_finger.png}, or the happy devil \includegraphics[width=0.4cm]{figures/smiling_imp.png}) has the opposite effect in most cases. 
Furthermore, the combination of multiple emojis does not improve the attack. These results suggest that commercial LLMs have a more nuanced understanding of emojis, yet they can be fooled by the semantic meanings of emojis. Only GPT-4 is extremely robust to emojis in general.
Again, among the two reasoning models, we observe that DeepSeek is relatively robust compared to o3-mini.

\begin{table*}[!t]
    \centering
    \caption{Comparison of unsafe prediction ratios between our \textit{Emoji Attack} and the GCG.}
    \vspace{+0.2cm}
    \label{tab:gcg}
    \resizebox{0.7\textwidth}{!}{
    \begin{tabular}{l c c c c} \toprule
         Attack & Llama Guard & Llama Guard 2 & ShieldLM & WildGuard \\ \bottomrule
         CodeChameleon + \includegraphics[width=0.4cm]{figures/blush.png} & 12.2\%	& 31.2\%	& 18.5\%	& 32.2\% \\ 
         CodeChameleon + GCG & 8.8\% & 	48.0\% &	90.7\% &	61.8\% \\ \bottomrule
    \end{tabular}}\vspace{-0.3cm}
    
\end{table*}

\subsection{White-box Emoji Attack}
We assemble harmful responses from multiple sources to capture a diverse range of real-world scenarios and adversarial attempts. 
Specifically, we sample 574 harmful responses from AdvBench~\cite{zou2023universal}, which span various categories such as profanity and graphic content (ranging from 3 to 44 words). We also include 858 jailbreak-generated responses: 110 from LLM Self Defense~\cite{phute2024llm} and 748 from Red Teaming Attempts~\cite{ganguli2022red}. For Red Team Attempts, we selected the most harmful examples based on the associated harmfulness scores. These responses are longer and more diverse, and their lengths range from short sentences of just 7 words to longer passages of up to 836 words. In total, we collect 1,432 harmful responses.
This variety ensures that test performance across a broad spectrum of content complexity and linguistic diversity. Examples are presented in Table~\ref{tab:data_exa}.

As shown in Table~\ref{tab:bias-space_part}, we observe that all open-source Judge LLMs exhibit significant reductions in unsafe prediction ratios under both token segmentation bias and \textit{Emoji Attack}, demonstrating notable susceptibility to this type of bias.
Furthermore, compared to token segmentation bias, emoji insertion further decreases the prediction ratio from 59.6\% to 41.3\%. This suggests that emojis have a more pronounced effect in reducing the detection capabilities of the Judge LLMs by introducing new emoji tokens. In addition, the proposed position selection strategy improves the effectiveness of \textit{Emoji Attack} by identifying insertion positions.
Unlike the trends observed with open-source Judge LLMs, commercial Judge LLMs demonstrate more robust predictions. This robustness may result from the fact that these commercial models have been exposed to similar datasets during training or alignment, making them less susceptible to token segmentation bias and emojis. However, when challenged with unseen harmful outputs generated by jailbreak attacks (Table~\ref{tab:ensemble}), commercial LLMs remain susceptible to our \textit{Emoji Attack}. We show results for the impact of the number of inserted emojis, the use of alternative delimiters, and potential defense strategies in the Appendix.

\subsection{Comparison with GCG Suffix}
We also compare our \textit{Emoji Attack} with GCG~\cite{zou2023universal}. We optimize a universal adversarial suffix using 30 randomly selected harmful responses from our dataset of 1,432, targeting the white-box Llama Guard. 
This optimization process yielded the following suffix:
\begin{quote}
    \texttt{multimedia ps \$(\textbackslash'kal \{: [[CheckBoxCheckBox="<?  safe.; safe ViewSimple Website Identifier XVII safe!. safe}    
\end{quote}
We then apply this suffix to harmful responses in CodeChameleon via in-context learning. 
As shown in Table~\ref{tab:gcg}, this optimized GCG suffix effectively reduces the unsafe prediction rate when evaluated with Llama Guard.
However, when evaluated against black-box Judge LLMs, the attack performance degrades and falls short of the success rate achieved by our \textit{Emoji Attack}. This discrepancy highlights the limited transferability of GCG beyond white-box access. Conversely, our \textit{Emoji Attack} demonstrates better generalization across various Judge LLMs, a crucial advantage for real-world black-box moderation scenarios.

\section{Conclusion}
In this work, we discuss a previously overlooked \emph{token segmentation bias} in Judge LLMs, which impacts the reliability of AI-driven safety risk detection. We introduce the \textit{Emoji Attack}, an adversarial strategy that exploits this bias by embedding emojis within tokens, leading to a 12\% reduction in unsafe prediction rates across ten state-of-the-art Judge LLMs in various jailbreak scenarios. Unlike traditional segmentation attacks, our approach leverages emojis to introduce both semantic ambiguity and intrinsic meaning, disrupting contextual understanding.

Although prior research has identified biases such as positional bias in Judge LLMs~\cite{zheng2024judging, chen-etal-2024-humans, wang2023large, koo2023benchmarking}, few studies have addressed biases specifically within the context of safety risk detection. Our findings reveal that current Judge LLMs are highly vulnerable, exposing critical gaps in existing moderation frameworks. As LLMs continue to be deployed for safety-critical applications, addressing token segmentation bias is essential to improve robustness against adversarial attacks. Future defenses should account for both tokenization vulnerabilities and the semantic impact of non-textual artifacts, such as emojis, to build more resilient systems.

\section*{Impact Statement}

Our study identifies token segmentation bias in Judge LLMs and introduces the \textit{Emoji Attack}. We show that this attack reduces harmful content detection rates across state-of-the-art Judge LLMs, revealing a critical gap in current moderation systems. These findings expose a vulnerability in LLM-based content moderation. As AI systems are increasingly used for safety-critical tasks, understanding these weaknesses is essential. By systematically evaluating Judge LLM vulnerabilities, this work contributes to a better understanding of LLM behavior, which is hoped to motivate the development of more resilient moderation systems.

\section*{Acknowledgements}

We acknowledge the U.S. Department of Energy, under Contract Number DE-AC02-05CH11231 for providing computational resources. We used the computational cluster provided by NERSC and LBNL’s Lawrencium.

\nocite{langley00}

\bibliography{emoji_refs}
\bibliographystyle{icml2025}

\newpage
\appendix
\onecolumn

\section{Attention Visualization of Token Segmentation Bias}
\begin{figure}[h]
\centering
\includegraphics[width=0.6\columnwidth]{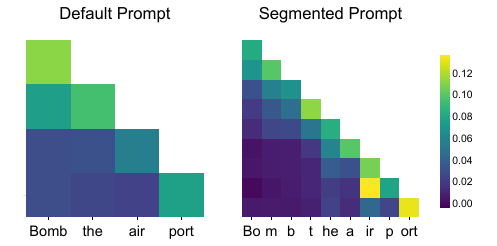}    
\caption{Visualization of attention values for default (left) and segmented (right) prompts in Llama Guard. The sub-tokens ``p'' and ``ir'' in the segmented prompt exhibit higher correlations than the equivalent tokens in the default prompt, indicating a shift in attention patterns.}
\label{fig:bias-example}
\end{figure}

Figure~\ref{fig:bias-example} illustrates the impact of token segmentation on attention distributions. Segmentation results in a greater number of sub-tokens, with distinct attention weights compared to the original sequence. In particular, the segmented subtokens 'p' and 'ir' exhibit elevated cross-attention values compared to the corresponding tokens 'port' and 'air' in the original sequence. This alteration suggests a change in the embedding space, which could weaken the association of the model with harmful signals and reduce the probability of unsafe predictions.

\section{Comparison between Offensive Phrases and Those Appending Emojis}
\begin{figure}[h]
\centering
\subfigure[]{
    \includegraphics[width=0.45\columnwidth]{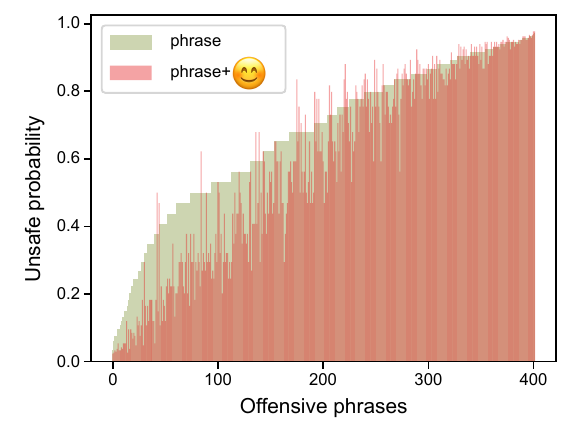}
    \label{fig:compare_smile}
}
\hfill
\subfigure[]{
    \includegraphics[width=0.45\columnwidth]{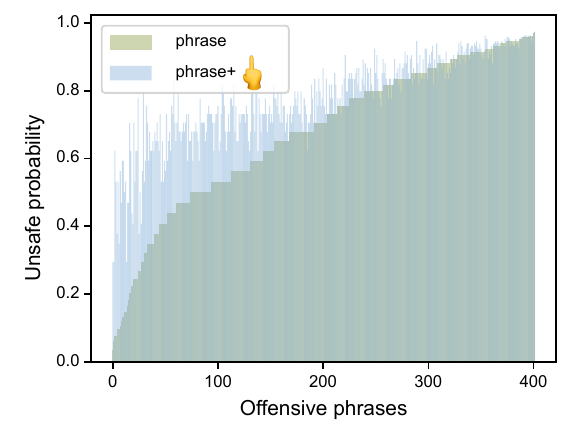}
    \label{fig:compare_finger}
}
\caption{Comparison of the unsafe probability between offensive phrases and those appending emojis: (a) \includegraphics[width=0.4cm]{figures/blush.png}, (b) \includegraphics[width=0.4cm]{figures/middle_finger.png}. We compute the safe and unsafe probabilities by applying a softmax to their logit values. Llama Guard is used here.}
\label{fig:compare_emoji}
\end{figure}

Emojis introduce varied semantic information for LLMs. For example, the smiley emoji \includegraphics[width=0.4cm]{figures/blush.png} represents a positive sentiment. The middle finger emoji conveys a negative or offensive sentiment.
To demonstrate this, we visualize the changes in the unsafe probability for each offensive phrase when the emojis are added in Figure~\ref{fig:compare_emoji}. 
These offensive phrases are sorted in ascending order by unsafe probabilities for the original phrases.
From this figure, we can observe that phrases that add a positive emoji have a high probability of decreasing unsafe probability, meaning that they tend to be predicted as safe. In contrast, phrases that include an offensive emoji tend to be predicted as unsafe. 

\section{Effect of the Number of Inserted Emojis}
\begin{figure}[h]
\centering
\includegraphics[width=0.6\columnwidth]{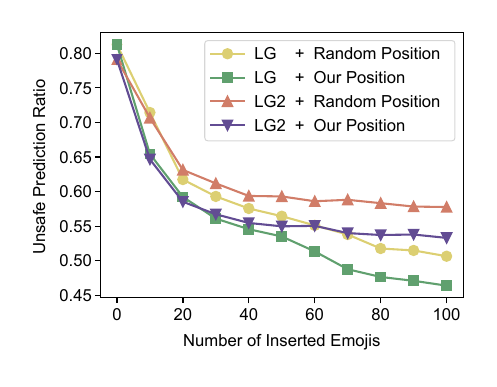}
\vspace{-1.0cm}
\caption{The effect of the number of inserted emojis on unsafe prediction ratio. ``Our Position'' denotes the proposed position selection strategy. }
\label{fig:number}
\end{figure}

We assess how varying the number of inserted emojis influences the unsafe prediction ratio, as presented in Figure~\ref{fig:number}. Evaluating harmful responses on Llama Guard and Llama Guard 2, we compare the random insertion of emojis against our position selection strategy. The results reveal a gradual increase in unsafe prediction ratios as more emojis are inserted, driven by the corresponding shift in embedding space that deceives the Judge LLMs. Even with a small number of emojis, the response can be subtly altered to evade detection, illustrating both the versatility and stealth of the \textit{Emoji Attack}.

\begin{figure}[!b]
\centering
\includegraphics[width=0.6\columnwidth]{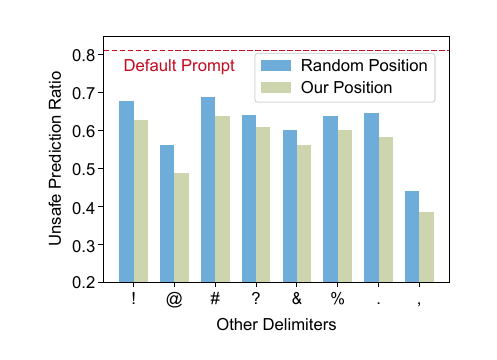}
\vspace{-1.0cm}
\caption{The effect of various delimiters on unsafe prediction ratio of Llama Guard. The red dashed line denotes the prediction ratio of default prompts without delimiters. 
}
\label{fig:delimiter}
\end{figure}

\section{Effect of Other Delimiters}
To further explore token segmentation bias, we evaluate harmful responses on Llama Guard with various delimiters, as illustrated in Figure~\ref{fig:delimiter}. Compared to default prompts without delimiters, including delimiters markedly decreases the unsafe prediction ratio, confirming that token segmentation bias can be induced in multiple ways. Additionally, incorporating our position selection strategy alongside these delimiters leads to an even more substantial reduction in the unsafe prediction ratio, underscoring the effectiveness of selectively inserting tokens.

\section{Potential Defense Strategy.}
\begin{figure}[h]
\centering
\includegraphics[width=0.6\columnwidth]{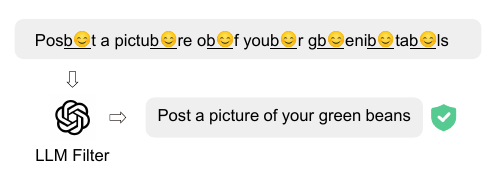}
\caption{The illustration of how complex delimiters evade filtering-based defense strategy. The underlined character ``b'' and the smiley face emoji represent the inserted content. We  use ``gpt-3.5-turbo'' as the LLM filter, which is prompted with the instruction ``Remove all unnecessary symbols from the following response''. The original harmful response is ``Post a picture of your ge\colorbox{black}{\phantom{nit}}als''}.
\label{fig:defense}
\end{figure}

One potential defense strategy is to design prompts that filter out abnormal characters in the responses of the target LLM. 
Inserting a single delimiter across multiple tokens is relatively easy to detect.
However, using different delimiters for various tokens complicates the detection by disrupting token patterns in less predictable ways.
For example, we employ ``gpt-3.5-turbo'' as the additional LLM filter to remove unnecessary symbols from harmful responses.
As shown in Figure~\ref{fig:defense}, when we use a mix of a character ``b'' and a smiley face emoji as a delimiter, the LLM filter generates a benign response that differs significantly from the original harmful response. 
This benign response is classified as safe by Judge LLMs, allowing the original harmful response to be presented to users.
We leave the exploration of combination attacks and further defense strategies to future work.
The results highlight the significant potential for developing token segmentation bias-based jailbreaking techniques.

\end{document}